\documentclass[runningheads]{llncs}
\usepackage{jj}
\usepackage[T1]{fontenc}
\begin{document}
\setlength{\belowdisplayskip}{5pt}
\setlength{\abovedisplayskip}{5pt}

\title{From Pixels to Torques with Linear Feedback}
%
%
\author{Jeong Hun Lee\inst{1} \and
Sam Schoedel \inst{1} \and
Aditya Bhardwaj\inst{2} \and
Zachary Manchester \inst{1}}
\authorrunning{J. Lee et al.}
%
\institute{Robotics Institute, Carnegie Mellon University, Pittsburgh 15213, USA
\email{\{jeonghunlee, sschoedel, zacm\}@cmu.edu}\\ \and
University of Chicago, Chicago 60637, USA\\
\email{a7b@uchicago.edu}}

\maketitle              
\begin{abstract}
We demonstrate the effectiveness of simple observer-based linear feedback policies for ``pixels-to-torques'' control of robotic systems using only a robot-facing camera. Specifically, we show that the matrices of an image-based Luenberger observer (linear state estimator) for a ``student'' output-feedback policy can be learned from demonstration data provided by a ``teacher'' state-feedback policy via simple linear-least-squares regression. The resulting linear output-feedback controller maps directly from high-dimensional raw images to torques while being amenable to the rich set of analytical tools from linear systems theory, allowing us to enforce closed-loop stability constraints in the learning problem. We also investigate a nonlinear extension of the method via the Koopman embedding. Finally, we demonstrate the surprising effectiveness of linear pixels-to-torques policies on a cartpole system, both in simulation and on real hardware. The policy successfully executes both stabilizing and swing-up trajectory-tracking tasks using only camera feedback while subject to model mismatch, process and sensor noise, perturbations, and occlusions. Open-source code for all experiments can be found here: \url{https://roboticexplorationlab.org/projects/linear_pixels_to_torques.html}

\end{abstract}
%
%
%
\section{Introduction}

\begin{figure*}[t]
  \centering
  \includegraphics[height=4.5cm]{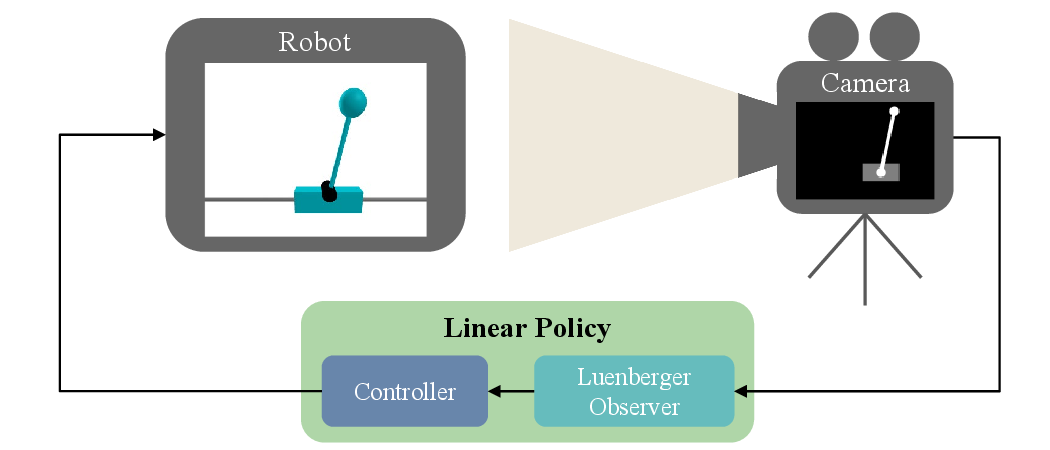}
  \caption{An overview of the problem we address, in which a simple linear output-feedback policy must control a robotic system using only feedback from images. In this work, the robotic system is both a simulated and real-world cartpole.}
  \label{fig:problem_setup}
  \vspace{-1\baselineskip}
\end{figure*}

Both model-based~\cite{xing_autonomous_2023, xiao_deep_2023, lyu_task-oriented_2023, laferriere_deep_2021, zhang_robust_2021, lu_image-based_2021, villarreal_mpc-based_2020, larsen_industrial_2006, wilson_relative_1996} and learning-based~\cite{quillen_deep_2018, mnih_playing_2013, pathak_curiosity-driven_2017, levine_end--end_2016, shi_waypoint-based_2023, wu_example-driven_2022, agarwal_legged_2023} control policies have demonstrated impressive results in robotics using vision-based sensory feedback. However, both suffer from inherent drawbacks: neural-network policies trained from images can require large amounts of data, which often must be generated in simulation before being transferred to a real robot. Meanwhile, model-based policies require separate machine-vision algorithms to extract features for downstream state estimation or model learning. Recently, data-driven switched linear control policies have been applied directly to image feedback with promising results~\cite{suh_surprising_2021}. However, this was limited to a quasi-static setting. We aim to apply data-driven linear feedback policies to the control of dynamic robotic systems directly from images as shown in Fig.~\ref{fig:problem_setup}.

We demonstrate that it is possible to perform ``pixels-to-torques'' control of robotic systems with simple observer-based linear feedback policies in both simulation and on real hardware. Specifically, we show that an image-based linear output-feedback policy can be designed by learning a Luenberger observer (linear state estimator) via linear-least-squares (LLS) regression over supervised demonstration data. The result is a pixels-to-torques policy that is interpretable and can be analyzed using the rich set of tools from classical linear systems theory, all while inferring states directly from high-dimensional images. We leverage this interpretability to promote stability of the closed-loop policy via a convex extension of the LLS problem. The resulting linear output-feedback policy can achieve good performance while  learning from small amounts of data, which we demonstrate on a classic cartpole system. Specifically, the policy is able to avoid sim-to-real transfer issues by learning directly from hardware data, and it successfully stabilizes the system while being robust to perturbations, unmodeled dynamics, and visual occlusions.
To summarize, our contributions are:
\begin{itemize}
  \item A data-efficient ``student-teacher'' methodology for learning linear, visual output-feedback policies.
  \item A convex formulation of the least-squares observer-learning problem that includes closed-loop stability constraints.
  \item A nonlinear extension of our method based on the Koopman embedding.
  \item Demonstration of linear ``pixels-to-torques'' control on a real-world cartpole.
\end{itemize}

The remainder of the paper is organized as follows: In Section \ref{sec:relatedworks}, we provide a broad survey of related works on image-based control and estimation in robotics. In Section \ref{sec:background}, we provide an overview of topics from linear control, state estimation, and Koopman-operator theory. In Section \ref{sec:methodology}, we describe our linear policy framework as well as the student-teacher method developed to learn a Luenberger observer. We also introduce a Koopman-based extension for nonlinear control settings. Section \ref{sec:results} presents experimental results demonstrating the method's performance on a cartpole system both in simulation and on real-world hardware. Lastly, we summarize our conclusions in Section \ref{sec:conclusion}.

\section{Related Works}\label{sec:relatedworks}

\subsection{Learning Dynamics Models from Images}

Over the last decade, the deep-learning community developed dynamics-model learning from images for model-based reinforcement learning (RL) and model-predictive control (MPC)~\cite{ye_object-centric_2020, fragkiadaki_learning_2016, hafner_learning_2019, finn_deep_2016, watter_embed_2015, finn_deep_2017}. To avoid the high-dimensionality of pixel space, the dynamics were initially learned over lower-dimensional, object-centric states to explicitly track objects in the image scene~\cite{ye_object-centric_2020, fragkiadaki_learning_2016}. While the states in these works were explicitly predefined or incorporated into a known problem structure, subsequent work also learned the lower-dimensional latent state space for robotics settings, which include simulated gym environments~\cite{hafner_learning_2019, watter_embed_2015} and real-world object manipulation~\cite{finn_deep_2016}. Specifically, Finn et. al.~\cite{finn_deep_2016} and Watter et. al.~\cite{watter_embed_2015} learned latent spaces where the learned dynamics were also locally linear and time varying. This allowed for simple stochastic policies to be effectively used for downstream control tasks. In contrast to object-centric models, Finn et. al.~\cite{finn_deep_2017} also learned image-to-image dynamics (termed ``visual foresight'') and coupled it with a stochastic MPC controller for pushing manipulation tasks.

Recently, ``deep'' Koopman methods have used deep neural networks to \emph{lift} observations into a higher-dimensional latent space where the dynamics behave linearly~\cite{lusch_deep_2018, xiao_deep_2023, lyu_task-oriented_2023, laferriere_deep_2021}. Laferriere et. al~\cite{laferriere_deep_2021} and Xiao et. al~\cite{xiao_deep_2023} extend deep Koopman to pixel-to-control tasks for robotics, including cartpole stabilization. Specifically, images were first embedded to lower-dimensional latent spaces using autoencoders before being lifted via multi-layer perceptrons (MLPs). Lyu et. al.~\cite{lyu_task-oriented_2023} expands upon this further to use constrastive encoders for cartpole-swingup tasks. We note the similarities of deep Koopman to \emph{time-varying}-latent-space-dynamics learning by Finn et. al.~\cite{finn_deep_2016} and Watter et. al.~\cite{watter_embed_2015}. While both methods aim to learn dynamics models for downstream controllers, our method aims to learn a linear state estimator for which images are observations.

\subsection{Learning Policies with Image Feedback}

Learning control policies directly from visual inputs has been a widely used practice in the deep-RL community ~\cite{mnih_playing_2013, pathak_curiosity-driven_2017, agarwal_legged_2023, levine_end--end_2016}. While initially performed within virtual environments~\cite{mnih_playing_2013, pathak_curiosity-driven_2017}, deep-neural-network image-feedback policies have quickly been extended to real-world robotics tasks, including manipulation via diffusion policies~\cite{chi_diffusion_2024} and quadruped locomotion over various terrains with egocentric vision~\cite{agarwal_legged_2023}. These methods, while demonstrating impressive results, are limited by the need for large amounts of training data. To scale to learning on real-world hardware, Levine et. al.~\cite{levine_end--end_2016} proposed a guided policy search method for efficiently learning an end-to-end pixels-to-torques policy for manipulation tasks. Specifically, ``guiding'' linear controllers with privileged full-state feedback are optimized and used to provide supervised training data for learning the pixels-to-torques policy. Rather than learning a deep end-to-end policy, our method learns a simple Luenberger observer from the guided supervised data to achieve image-to-state estimation for a downstream state-feedback controller.

\subsection{Vision-Based Pose and State Estimation}

Deep learning has been widely applied to vision-based pose and state estimation across many robotics applications~\cite{xing_autonomous_2023, richter_autonomous_2021, lee_camera--robot_2020, yoo_toward_2023, zuo_craves_2019, lambrecht_towards_2019}. Applications include both external feature detection for a downstream controller to track~\cite{richter_autonomous_2021, xing_autonomous_2023} and estimation of the robot's own pose or state~\cite{zuo_craves_2019, lambrecht_towards_2019, lee_camera--robot_2020, yoo_toward_2023}. A commonality across these learning-based methods is the need for large amounts of data, necessitating data synthesis and sim-to-real transfer. As a result, Lu et. al. looked to differentiable rendering for both robot pose and state estimation from images~\cite{lu_tracking_2023, lu_image-based_2023}. Specifically, Lu et. al. rendered predicted images of the robot and calculated the corresponding Jacobians for an extended Kalman filter, resulting in online robot-state estimation from a single camera~\cite{lu_tracking_2023}. However, this method requires a differentiable rendering model of the robot to generate predicted observations. Therefore, we aim to determine if it's possible to directly learn an image-based Luenberger observer instead of relying on an explicit observation model.

\subsection{Effectiveness of Linearization}

Over many decades, linear controllers and state estimators have proven to be powerful tools within the controls community thanks to their simplicity, ease of analysis, and often surprisingly good performance on nonlinear systems~\cite{kalman_new_1960, kailath_linear_1980}. Recently, under the label of state-space models (SSMs), linear models have also become popular in the deep-learning community~\cite{gu_mamba_2023, gu_efficiently_2022, gu_hippo_2020}. Specifically, Gu et. al.~\cite{gu_mamba_2023, gu_efficiently_2022} demonstrated that structured linear models in the latent spaces of deep networks can model long-range dependencies found in many applications like speech recognition, language processing, and genomics~\cite{gu_hippo_2020}. The most recent implementations have produced performance matching or exceeding that of widely-adopted transformers while being much more computationally efficient~\cite{gu_mamba_2023}. We note the strong similarity of SSMs to deep-Koopman methods~\cite{lusch_deep_2018}, specifically in their learning of linear dynamics in a latent space.

In addition to deep-learning methods, non-neural linear alternatives have also been shown to be surprisingly effective in robotics~\cite{haggerty_control_2023, bruder_data-driven_2021, lu_tracking_2023} and vision-based applications~\cite{suh_surprising_2021, lu_image-based_2021}. By applying approximate Koopman operators~\cite{korda_linear_2018}, both Haggerty et. al.~\cite{haggerty_control_2023} and Bruder et. al.~\cite{bruder_data-driven_2021} were able to model highly nonlinear, high-dimensional soft manipulators with no neural components. Within the context of quasi-static pile manipulation, Suh et. al.~\cite{suh_surprising_2021} showed that switched linear visual-foresight models can outperform deep counterparts in both prediction error and closed-loop control performance. However, applying these model-learning methods directly to image-based observations of open-loop unstable dynamical systems can be difficult. Koopman liftings of high-dimensional images are computationally expensive, and linear visual-foresight models have been restricted to quasi-static systems. Therefore, we aim to determine if it is possible to avoid the development of a Koopman dynamics model by directly learning an observer-based linear feedback policy.

\section{Background}\label{sec:background}

This section provides a brief review of linear control, state estimation, and Koopman-operator theory. We refer the reader to the substantial existing literature on these topics for further details~\cite{brunton_modern_2021, mezic_koopman_2021, korda_linear_2018, kailath_linear_1980, bertsekas_stochastic_1996, bertsekas_reinforcement_2019}.

\subsection{Linear Control}

We aim to describe our controlled robot with a discrete-time linear time-invariant dynamics model,
\begin{equation}
    x_{k+1} = Ax_k + Bu_k, \label{eq:linear_dynamics_model}
\end{equation}
where $x_{k+1}, x_k \in \R^{n}$ are the robot states at time steps $k+1$ and $k$ respectively; $u_k \in \R^{m}$ are the control inputs (i.e., actions) at time step $k$; and $A \in \R^{n \times n}$, $B \in \R^{n \times m}$ are the dynamics coefficient matrices.

Using the dynamics described by \eqref{eq:linear_dynamics_model}, we can design a linear state-feedback controller to drive the robot state to the origin: 
\begin{equation}
    u_{k} = -Kx_{k}, \label{eq:linear_feedback_control}
\end{equation}
where $K \in \R^{m \times n}$ is the controller gain matrix. As a particular example, K could be determined by solving an LQR problem using the Riccati equation~\cite{kailath_linear_1980}.

To evaluate the stability of the controller, we substitute \eqref{eq:linear_feedback_control} into \eqref{eq:linear_dynamics_model} to obtain an expression for the closed-loop dynamics:
\begin{equation}
    x_{k+1} = (A-BK)x_k, \label{eq:closed_loop_dynamics}
\end{equation}
where we can see that $x_{k}$ will converge to the origin if $A-BK$ is stable:
\begin{equation}
    |\text{eigvals}[A-BK]| < 1, \label{eq:stable_closed_loop_dynamics_controller}
\end{equation}
where $|\text{eigvals}[A-BK]|$ are the magnitudes of the eigenvalues of the closed-loop dynamics.

\subsection{Linear State Estimation}

As seen in \eqref{eq:linear_feedback_control}, the implementation of a linear state-feedback controller requires knowledge of the full state, $x_k$. However, this is rarely the case; we usually only have access to information from the robot's sensors. For now, we will assume that the sensors only provide partial state information described by the following observation model:
\begin{equation}
    y_{k} = Cx_{k}, \label{eq:linear_observation_model}
\end{equation}
where $y_{k} \in \R^{l}$ are the observations (e.g., sensor readings, images, etc.) at time step $k$ and $C \in \R^{l \times n}$ is the observation matrix. 

An observer can then be designed to first predict states using the dynamics, \eqref{eq:linear_dynamics_model}, before correcting the predictions using sensor observations to provide the final full-state estimates. This is the core idea behind a Luenberger observer (linear state estimator):
\begin{equation}
    \hat{x}_{k+1} = \underbrace{A\hat{x}_{k} + Bu_{k}}_{\text{prediction}} + \underbrace{L(y_{k+1} - \hat{y}_{k+1})}_{\text{correction}}, \label{eq:luenberger_observer}
\end{equation}
where $\hat{x}_{k+1}$ is the \emph{estimated} state output by the observer at time step $k+1$; $\hat{y}_{k+1}$ are the corresponding \emph{predicted} observations; $y_{k+1}$ are the corresponding \emph{real-world} observations (i.e., sensor measurements); and $L \in \R^{n \times l}$ is the observer gain. $L$ is often designed with measurement noise and model uncertainty in mind, while the estimator's internal state $\hat{x}_{k}$ maintains a memory of aggregated information from previous time steps~\cite{kailath_linear_1980}. As an example, a Kalman filter can be used to design $L$ to optimally  estimate $\hat{x}_{k}$ when all uncertainties in the system correspond to additive Gaussian noise.

The state estimates, $\hat{x}_{k}$, from the Luenberger observer at time step $k$ can be subsequently passed to a linear state-feedback controller to calculate the corresponding control inputs,
\begin{equation}
    u_{k} = -K\hat{x}_{k}. \label{eq:estimated_linear_feedback_control}
\end{equation}
We refer to this combined controller-observer system as an \emph{output-feedback policy} with an \emph{internal} state, $\hat{x}$.

Similarly to $K$, $L$ can also be calculated to drive the state estimation errors, $x_{k} - \hat{x}_{k}$, asymptotically to zero for a given linear feedback controller. We first express $\hat{y}_{k+1}$ in terms of $\hat{x}_{k}$ and $u_{k}$ by substituting the prediction term of \eqref{eq:luenberger_observer} into \eqref{eq:linear_observation_model}:
\begin{equation}
    \hat{y}_{k+1} = C(A\hat{x}_{k} + Bu_{k}). \label{eq:predicted_observation}
\end{equation}
Then, we substitute \eqref{eq:estimated_linear_feedback_control} and \eqref{eq:predicted_observation} into  \eqref{eq:luenberger_observer} to further simplify the observer,
\begin{equation}
    \hat{x}_{k+1} = (I-LC)(A-BK)\hat{x}_{k} + Ly_{k+1}. \label{eq:luenberger_observer_complete_feedback}
\end{equation}
This allows us to analyze the stability of the Luenberger observer in a similar manner to the linear state-feedback controller as described by \eqref{eq:stable_closed_loop_dynamics_controller}. Specifically, we can express the condition for asymptotically converging state-estimation errors as:
\begin{equation}
    |\text{eigvals}[(I-LC)(A-BK)]| < 1. \label{eq:stable_closed_loop_dynamics_estimator}
\end{equation}
For clarity, we will define the matrix $A^{LK}$:
\begin{equation}
    A^{LK} = (I-LC)(A-BK). \label{eq:estimator_closed_loop_dynamics}
\end{equation}

\subsection{Koopman-Operator Theory}

Koopman-operator theory has seen increased adoption in robotics in recent years~\cite{xiao_deep_2023, haggerty_control_2023, laferriere_deep_2021, folkestad_koopman_2021, bruder_data-driven_2021, bruder_advantages_2021}. This is due to its ability to apply model-based linear (or bilinear) control directly to nonlinear systems, which we denote as
\begin{equation} \label{eq:discrete_dynamics} 
  x_{k+1} = f(x_k, u_k).
\end{equation} 
Assuming these dynamics are control affine, the nonlinear system can be represented \emph{exactly} by an infinite-dimensional bilinear system~\cite{surana_koopman_2016}. This bilinear Koopman model takes the form,
\begin{equation} \label{eq:bilinear_dynamics}
  z_{k+1} = A z_{k} + B u_{k} + \sum_{i=1}^{m} u_{k}^i C^i z_{k},
\end{equation}
where $z$ is the infinite-dimensional \emph{lifting} of the robot states, $x$. In practice, this is approximated by a finite-dimensional nonlinear mapping over candidate basis functions~\cite{haggerty_control_2023, folkestad_koopman_2021, bruder_advantages_2021} or deep neural networks~\cite{folkestad_koopnet_2022, laferriere_deep_2021, xiao_deep_2023}. We represent this finite-dimensional embedding as,
\begin{equation}
    z = \phi(x),
\end{equation}
where $z \in \R^{j}$. Likewise, $\phi$ is constructed in such a way that the ``unlifting'' is linear:
\begin{equation}
	x = G z.
\end{equation}

\begin{figure*}[t!]
  \centering
  \includegraphics[width=\textwidth]{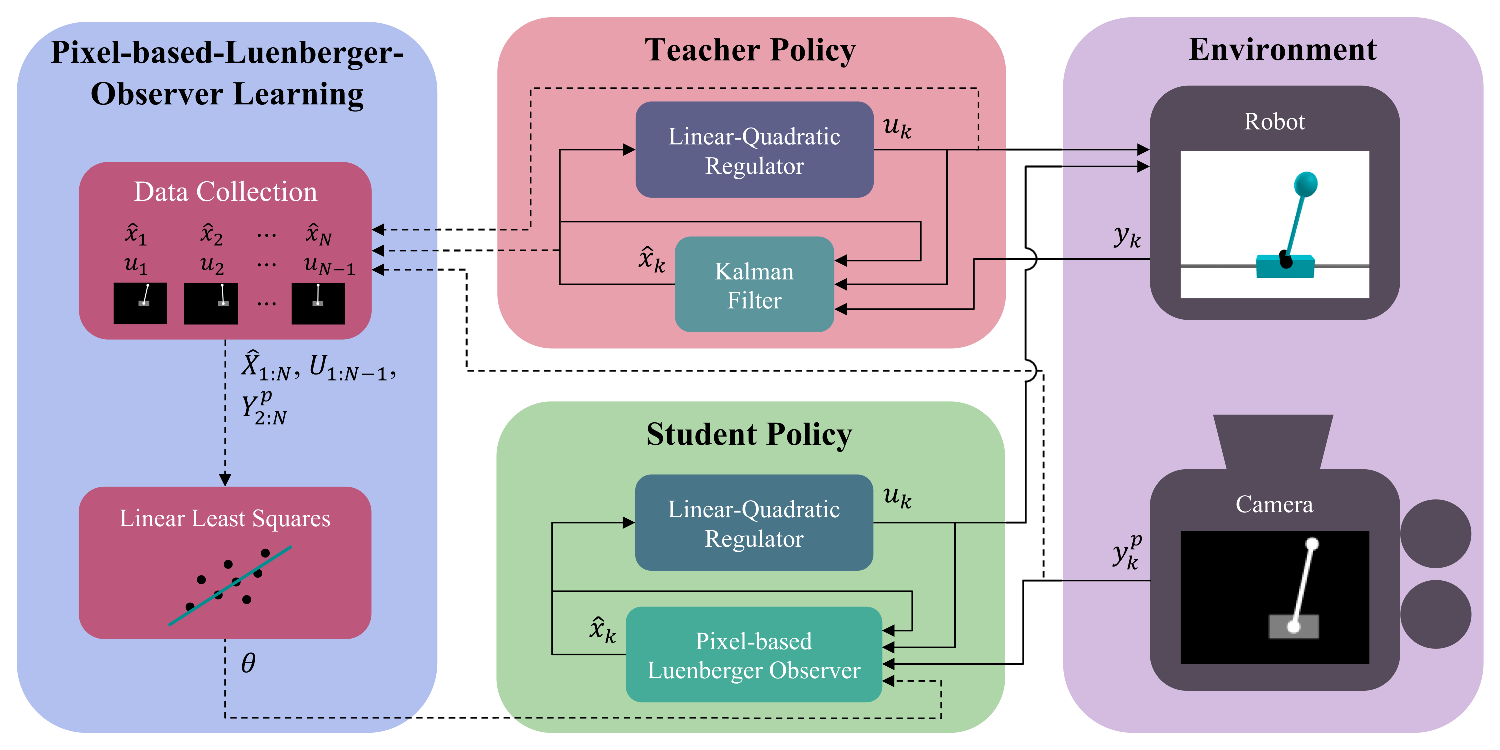}
  \caption{Overview of learning a linear output-feedback policy for pixels-to-torques control. In this work, a linear-quadratic-Gaussian (LQG) ``teacher'' policy is designed to control a cartpole system. Teacher demonstration data is collected as trajectories of the robot's estimated states, $\hat{X}_{1:N}$; control inputs, $U_{1:N-1}$; and corresponding images from a robot-facing camera, $Y^{p}_{2:N}$. Subsequent linear-least-squares (LLS) regression is performed over the data to determine the parameters, $\theta$, of a ``student'' policy's pixel-based Luenberger observer (linear state estimator). The student policy's controller can also be learned separately or cloned from the teacher's, as is the case in this work. The solid and dotted lines indicate processes that were performed online and offline respectively.}
  \label{fig:pixelol_overview}
  \vspace{-1\baselineskip}
\end{figure*}

\vspace{-1\baselineskip}

\section{Methodology}\label{sec:methodology}

For a robot controlled via image-based feedback, the development of an appropriate observation model, \eqref{eq:linear_observation_model}, requires a differentiable renderer~\cite{lu_image-based_2023} that we may not have access to. In addition, performing a gain-calculation procedure, such as solving the Riccati equation, may be too computationally expensive to perform using dynamics and observation models in pixel space~\cite{suh_surprising_2021}. Therefore, we directly learn the parameters of a linear feedback policy's Luenberger observer, which uses images from a robot-facing camera as sensory feedback. An overview of our method is shown in Fig.~\ref{fig:pixelol_overview}.

\subsection{Pixel-Based Luenberger Observer}

We begin by restating the Luenberger observer. Specifically, we substitute \eqref{eq:predicted_observation} into \eqref{eq:luenberger_observer} to express the Luenberger observer in terms of just the predicted robot states, $\hat{x}$; robot control inputs, $u$; and output observations in the form of high-dimensional, raw pixel values from a robot-facing camera, $y^{p}$:
\begin{equation}
    \hat{x}_{k+1} = \underbrace{(I-LC)A}_{A^{L}}\hat{x}_{k} + \underbrace{(I-LC)B}_{B^{L}}u_{k} + Ly^{p}_{k+1}, \label{eq:luenberger_observer_simplified}
\end{equation}
where $A^{L}$, $B^{L}$ denote new coefficient matrices.
We additionally introduce an affine term, $d \in \R^n$, to handle linearizations about non-zero goal points:
\begin{equation}
    \hat{x}_{k+1} = A^{L}\hat{x}_{k} + B^{L}u_{k} + Ly^{p}_{k+1} + d, \label{eq:luenberger_observer_simplified_with_affine}
\end{equation}
This yields the form of a pixel-based Luenberger observer whose coefficient matrices we will learn from subsequent supervised data. For simplification, we concatenate the matrices into
\begin{equation}
    \theta = \begin{bmatrix}
      A^{L} & B^{L} & L & d
    \end{bmatrix}. \label{concatenated_coeff_lo}
\end{equation}

\subsection{Learning a Pixel-Based Linear Output-Feedback Policy}

To gather supervised data, we first leverage a predesigned ``teacher'' policy as shown in Fig.~\ref{fig:pixelol_overview}. In our case, the teacher is a privileged linear-quadratic regulator (LQR) policy that utilizes the robot's built-in encoders. The teacher is used to collect supervised data in the form of demonstration trajectories of the predicted robot state, $\hat{X}_{1:N}$; control inputs, $U_{1:N-1}$; and pixel values from robot-facing-camera images,  $Y^{p}_{2:N}$, which are used as observations by a subsequently learned linear output-feedback policy for pixels-to-torques control. We refer to this learned policy as the ``student.''

The collected supervised trajectories are then concatenated and stored in the form of data matrices,
\begin{equation}
  W = \begin{bmatrix}
    \hat{X}_{1:N-1} \\
    U_{1:N-1} \\
    Y^{p}_{2:N} \\
    1
  \end{bmatrix} = \begin{bmatrix}
    \hat{x}_1   & \hat{x}_2   & \dots  & \hat{x}_{N-1}    \\
    u_1         & u_2         & \dots  & u_{N-1}          \\
    y^{p}_2     & y^{p}_3     & \dots  & y^{p}_{N}        \\
    1           & 1           & \dots  & 1
  \end{bmatrix}, \quad 
  \hat{X}_{2:N} = \begin{bmatrix}
    \hat{x}_2   & \hat{x}_3   & \dots  & \hat{x}_{N}
  \end{bmatrix}.
\end{equation}
This allows us to formulate the learning of the student's pixel-based Luenberger observer described by \eqref{eq:luenberger_observer_simplified_with_affine} as the following linear-least-squares (LLS) problem:
\begin{equation} \label{opt:pixelol}
  \underset{\theta}{\text{minimize}} \; \norm{\theta W - \hat{X}_{2:N}}_2^2.
\end{equation}
We additionally introduce sparsity-promoting $L_{1}$ regularization to reduce overfitting:
\begin{equation} \label{opt:pixelol_regularized}
  \underset{\theta}{\text{minimize}} \; \norm{\theta W - \hat{X}_{2:N}}_2^2 + \lambda ||L||_{1}.
\end{equation}

By combining the teacher's state-feedback controller with the newly designed Luenberger observer, the student output-feedback policy can perform closed-loop pixels-to-torques control as shown in Fig.~\ref{fig:pixelol_overview}. We note that regression can be separately performed over states and control histories to also learn a gain, $K$, for a new student controller. However, if the teacher is a linear feedback policy, this is equivalent to using the teacher's controller.

\subsection{Enforcing Stability with a Linear-Matrix-Inequality}

By leveraging existing techniques from linear systems theory~\cite{kailath_linear_1980, boyd_linear_1994}, we introduce a convex constraint to the observer-learning problem to ensure that state-estimation errors are guaranteed not to diverge. Specifically, we note that the stability condition described by \eqref{eq:stable_closed_loop_dynamics_estimator} is equivalent to the following spectral-norm condition,
\begin{equation}
    ||A^{LK}||_{2} < 1, \label{eq:spectral_norm}
\end{equation}
which can be rewritten as the following linear-matrix-inequality (LMI)~\cite{boyd_linear_1994}:
\begin{equation}
    \begin{bmatrix}
        I & A^{LK} \\
        A^{LK} & I
    \end{bmatrix} \succ 0, \label{eq:stability_LMI}
\end{equation}
where $\succ$ indicates positive definiteness. Enforcing this constraint in the LLS problem requires privileged information of the teacher's linear controller gain, $K$, which we assume access to. Combining \eqref{opt:pixelol_regularized} with \eqref{eq:stability_LMI} extends the original LLS problem into the following convex program, which we solve using the semidefinite programming solver COSMO~\cite{garstka_cosmo_2021}:
\begin{mini}|l|
  {\theta}{\norm{\theta W - \hat{X}_{2:N}}_2^2 + \lambda ||L||_{1},}{}{}
  \addConstraint{\begin{bmatrix}
        I & A^{LK} \\
        A^{LK} & I
    \end{bmatrix} \succ 0,}
\end{mini}
where $\theta$ and $W$ are now defined as
\begin{equation*}
  W = \begin{bmatrix}
    \hat{X}_{1:N-1} \\
    Y^{p}_{2:N} \\
    1
  \end{bmatrix}, \quad 
  \theta = \begin{bmatrix}
      A^{LK} & L & d
    \end{bmatrix}.
\end{equation*}

\subsection{Extension to Nonlinear Systems via Koopman Embedding}

For nonlinear control and estimation, the nonlinear dynamics can be linearized about a reference trajectory before being tracked by a time-varying linear controller (e.g., time-varying LQR) and estimator. However, doing so requires learning gains for every time step, $K_{k}$ and $L_{k}$, which can become computationally expensive and data-inefficient with high-dimensional image-based observations. Therefore, we introduce a Koopman-based extension to the pixel-based linear output-feedback policy that only requires learning a single Luenberger gain, $L$.

To do so, we start with a pre-specified choice of Koopman embedding and corresponding unlifting, 
\begin{align}
    z_{k} & = \phi(x_{k}), \label{eq:lifting} \\
    x_{k} & = Gz_{k}. \label{eq:unlifting}
\end{align}
We then replace the linear dynamics \eqref{eq:linear_dynamics_model} with the bilinear Koopman dynamics \eqref{eq:bilinear_dynamics} in our Luenberger Observer \eqref{eq:luenberger_observer}. Doing so yields the following Luenberger observer w.r.t. $z$:
\begin{equation}
    z_{k+1} = A z_{k} + B u_{k} + \sum_{i=1}^{m} u_{k}^i C^i z_{k} + L(y_{k} - \hat{y}_{k}), \label{eq:koopman_luenberger_observer}
\end{equation}
which can be further simplified in a similar manner to \eqref{eq:luenberger_observer_simplified}. An additional affine term can also be introduced, yielding:
\begin{equation}
    z_{k+1} = A^{L} z_{k} + B^{L} u_{k} + \sum_{i=1}^{m} u_{k}^i C^{L, i} z_{k} + Ly_{k+1} + d. \label{eq:koopman_luenberger_observer_simplified}
\end{equation}

We may now formulate a new LLS problem in the same manner as \eqref{opt:pixelol_regularized}:
\begin{equation} \label{opt:pixelol_koopman_regularized}
  \underset{\theta}{\text{minimize}} \; \norm{\theta W - \hat{Z}_{2:N}}_2^2 + \lambda ||\theta||_{1},
\end{equation}
where $\hat{Z}_{2:N}$ and $W$ are now defined as
\begin{align*}
  \hat{Z}_{2:N} & = \begin{bmatrix}
    \phi(\hat{x}_2)   & \phi(\hat{x}_3)   & \dots  & \phi(\hat{x}_{N})
  \end{bmatrix}, \\
  W & = \begin{bmatrix}
    \hat{Z}_{1:N-1} \\
    U_{1:N-1} \\
    (U^{1}\hat{Z})_{1:N-1} \\
    \vdots \\
    (U^{m}\hat{Z})_{1:N-1} \\
    Y^{p}_{2:N} \\
    1
  \end{bmatrix} = \begin{bmatrix}
    \hat{z}_1           & \hat{z}_2             & \dots     & \hat{z}_{N-1}    \\
    u_1                 & u_2                   & \dots     & u_{N-1}          \\
    u^{1}_{1}\hat{z}_1  & u^{1}_{2}\hat{z}_2    & \dots     & u^{1}_{N-1}\hat{z}_{N-1} \\
    \vdots              & \vdots                & \ddots{}  & \vdots \\
    u^{m}_{1}\hat{z}_1  & u^{m}_{2}\hat{z}_2    & \dots     & u^{m}_{N-1}\hat{z}_{N-1} \\
    y^{p}_2     & y^{p}_3     & \dots  & y^{p}_{N}        \\
    1           & 1           & \dots  & 1
  \end{bmatrix},
\end{align*}
and $\theta$ represents the concatenated matrices of the Koopman Luenberger observer:
\begin{equation*}
    \theta = \begin{bmatrix}
        A^L & B^L & C^{L, 1} & \dots & C^{L, m} & L & d
    \end{bmatrix}.
\end{equation*}
As before, this Koopman-based extension of the pixel-based Luenberger observer can be combined with the teacher's state-feedback controller (e.g., time-varying LQR) or a newly designed Koopman-based MPC controller~\cite{bruder_advantages_2021, folkestad_koopman_2021, korda_linear_2018}. The lifting and unlifting operations, \eqref{eq:lifting}-\eqref{eq:unlifting}, can be used to pass state information between a state-feedback controller and the Koopman-based Luenberger observer. This creates a pixels-to-torques policy that can now track trajectories for nonlinear systems.

\begin{figure*}[b!]
    \vspace{-1\baselineskip}
    \begin{subfigure}{\linewidth}
        \centering
        \includegraphics[height=4.25cm]{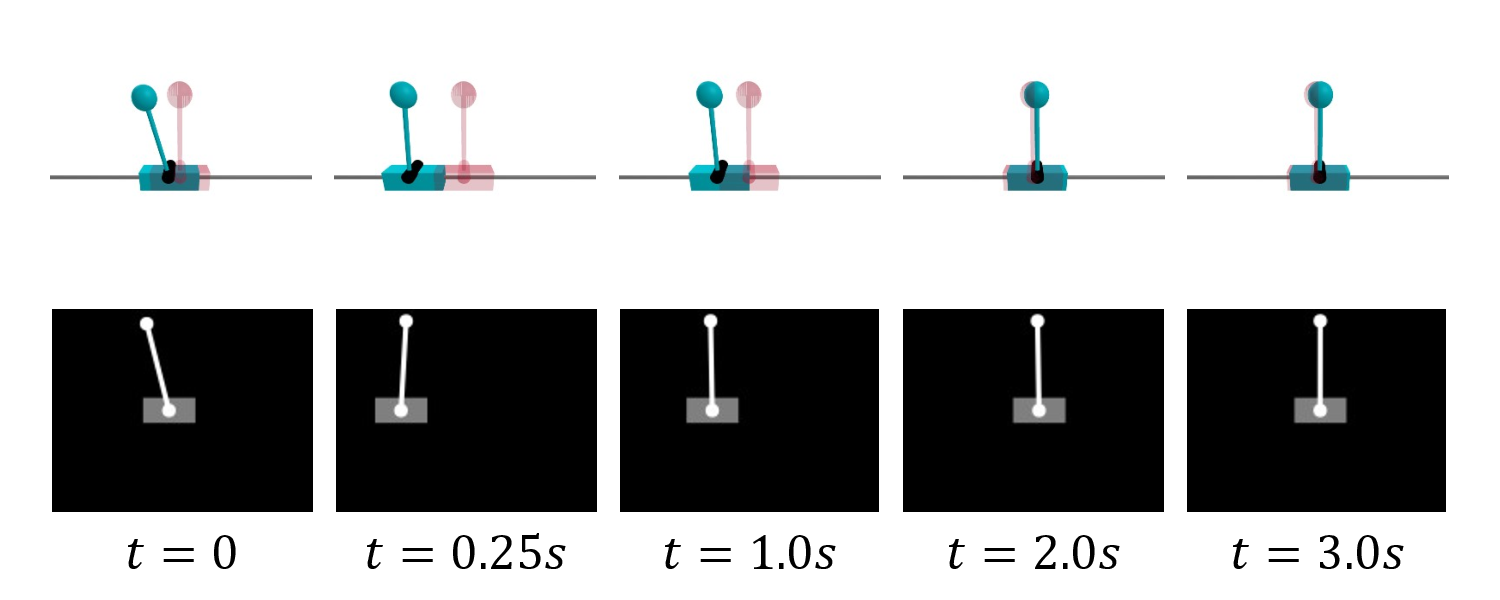}
        \caption{A visualization of a cartpole-stabilization trajectory executed by a pixel-based linear output-feedback policy. The top row depicts the simulated cartpole environment while the bottom row shows the corresponding rendered images (i.e., simulated camera) used by the pixels-to-torques policy. The goal configuration is shown in \color{red} red\color{black}.}
        \label{fig:stabilizing_trajectory}
    \end{subfigure}
    \par\medskip
    \begin{subfigure}{.49\linewidth}
        \centering
        \includegraphics[width=\linewidth, height=4cm]{fig3b_student_error_sweep_study.tikz}
        \caption{Stabilization-error performance}
        \label{fig:stabilization_error}
    \end{subfigure}%
    \hfill
    \begin{subfigure}{.49\linewidth}
        \centering
        \includegraphics[width=\linewidth, height=4cm]{fig3c_student_num_success_sweep_study.tikz}
        \caption{Success rate of stabilization}
        \label{fig:stabilization_success}
    \end{subfigure}\\
    \vspace{-0.5\baselineskip}
    \caption{Stabilization performance vs. training trajectories of a pixel-based linear output-feedback policy tasked with stabilizing a cartpole using a robot-facing camera. Test stabilizations from 100 different initial conditions were evaluated with stabilization error defined as the $L^2$ error of the final state w.r.t the upright goal state. The median error is shown as a thick line, while the shaded regions represent the 5\% to 95\% bounds. The corresponding success rate of stabilization from the 100 initial conditions is also shown.}
    \label{fig:stabilization_performance}
\end{figure*}


\section{Experimental Results}\label{sec:results}

\subsection{Simulation Setup}

\begin{figure*}[t!]
    \begin{subfigure}{.23\linewidth}
        \centering
        \includegraphics[height=2.2cm]{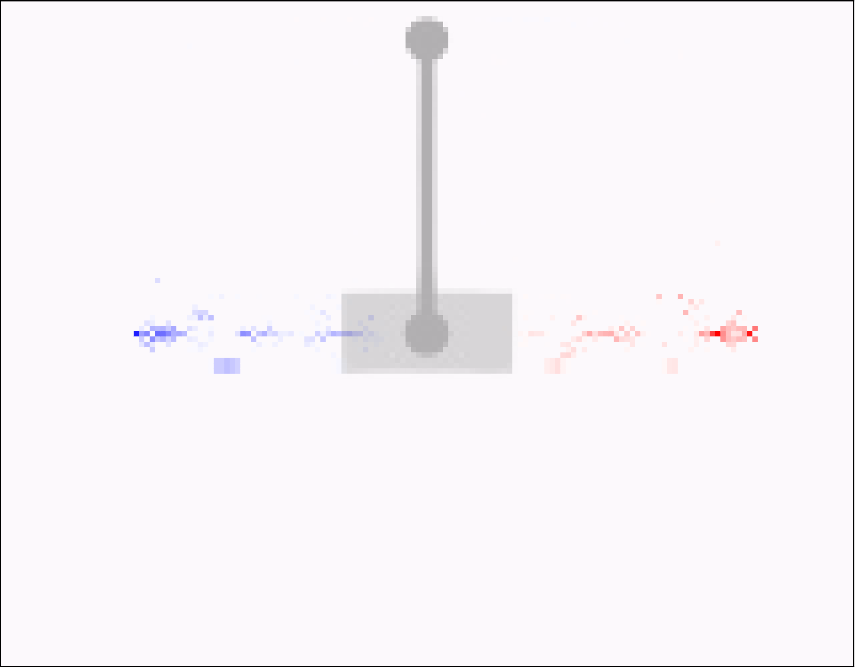}
        \caption{Cart position}
        \label{fig:position_gain}
    \end{subfigure}
    \hfill
    \begin{subfigure}{.23\linewidth}
        \centering
        \includegraphics[height=2.2cm]{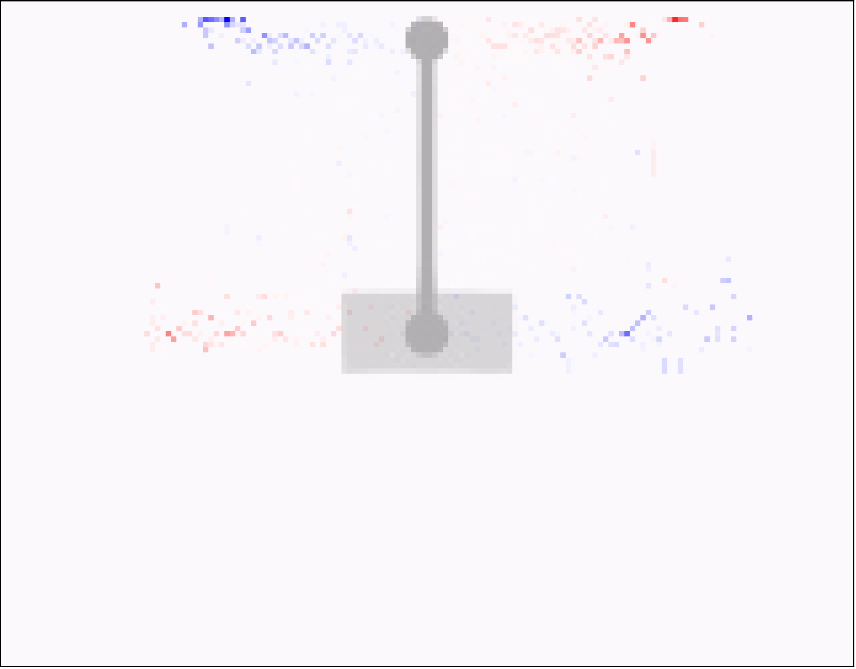}
        \caption{Cart velocity}
        \label{fig:angle_gain}
    \end{subfigure}
    \hfill
    \begin{subfigure}{.23\linewidth}
        \centering
        \includegraphics[height=2.2cm]{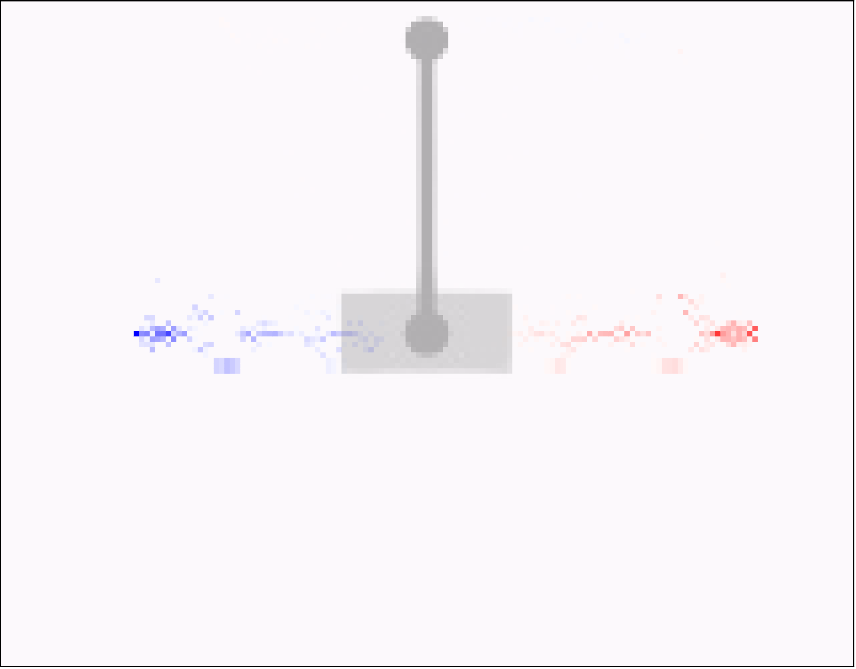}
        \caption{Pole angle}
        \label{fig:position_velocity_gain}
    \end{subfigure}
    \hfill
    \begin{subfigure}{.285\linewidth}
        \centering
        \includegraphics[height=2.2cm]{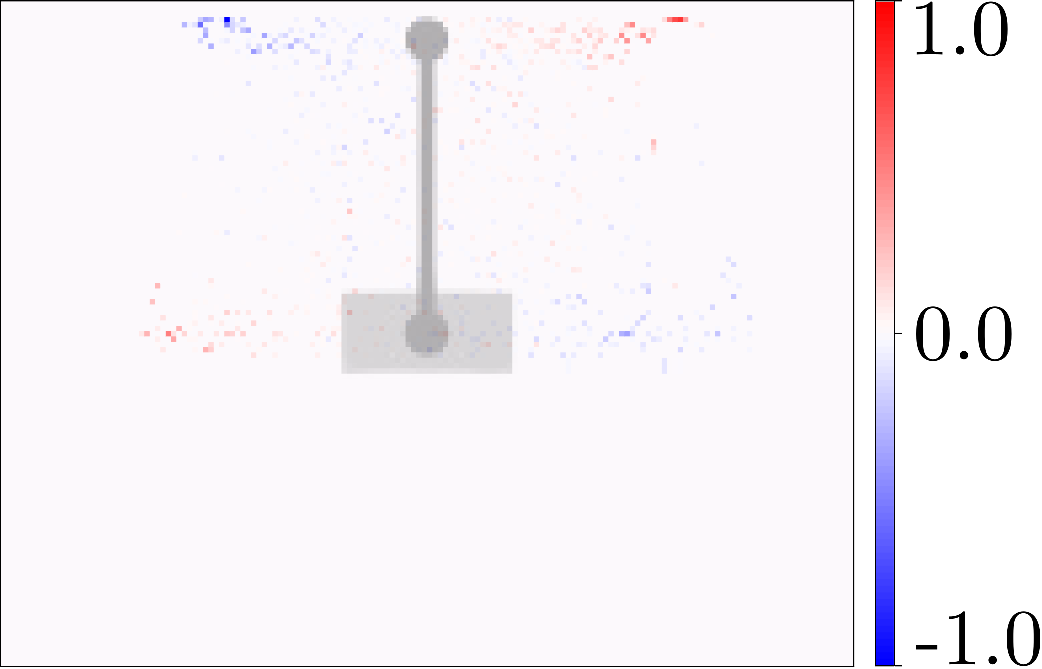}
        \caption{Pole angular velocity}
        \label{fig:angular_velocity_gain}
    \end{subfigure}\\
    \vspace{-0.5\baselineskip}
    \caption{Heat-map visualizations of each normalized row of the pixel-based Luenberger observer's gain matrix, $L$. A cartpole visualization is also overlayed for reference. Each row of $L$ corresponds to the correction an image observation contributes to a respective state variable. Interestingly, visual features can be distinguished for each state variable: the cart for the cart position with the addition of the pole tip for the pole angle. The velocities also have similar features.}
    \label{fig:observer_gain_visualization}
    \vspace{-1.0\baselineskip}
\end{figure*}

We evaluate the closed-loop performance of our pixel-based linear output-feedback policy in a simulated cartpole environment. We specify two models: a \emph{nominal} model, which is simplified and contains up to $5\%$ parametric model error, and a \emph{true} model, which is used exclusively for simulating the system in a ``real-world'' environment to evaluate the algorithm's performance. We additionally introduce Gaussian noise applied to the control inputs of the true model. The nominal model is used to design the teacher LQG policy, which uses encoder measurements of the cartpole's configuration as observations for its state estimator. Observation noise is simulated as encoder quantization in the true model. To simulate a camera sensor for the pixels-to-torques student policy, we use Makie~\cite{danisch_makiejl_2021} to render greyscale images of the simulated cartpole at a resolution of $125\times160$ pixels. To discern these image-based observations from the true, ``real-world'' system, we visualize the true system in Meshcat~\cite{deits_meshcatjl_2024}. The pixels-to-torques student policy is learned on a computer equipped with a $64$-core AMD Threadripper CPU and $64$GB of RAM.

\begin{figure*}[t!]
    \begin{subfigure}{\linewidth}
        \centering
        \includegraphics[width=\textwidth]{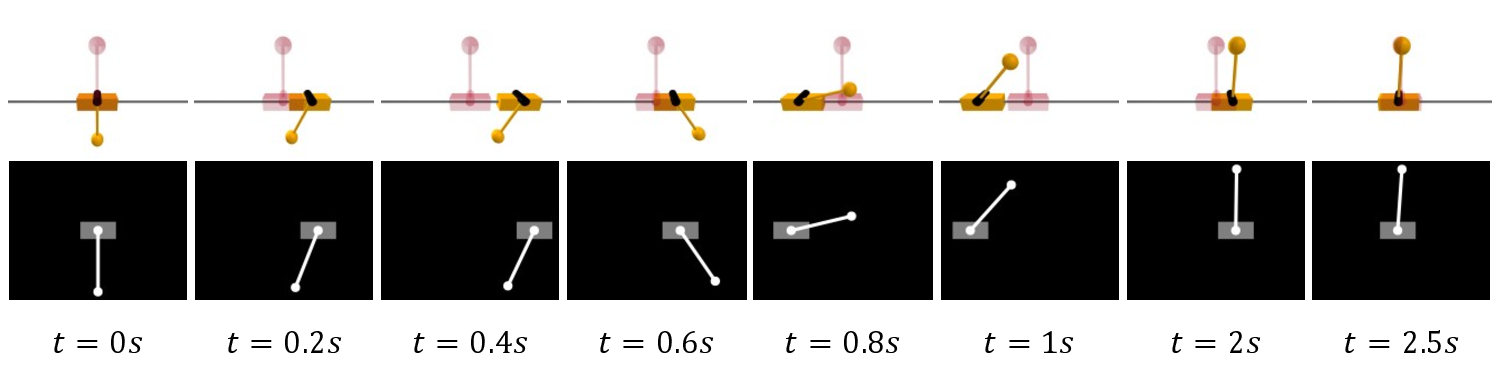}
        \caption{Cartpole swing-up trajectory visualization. The goal configuration is shown in \color{red} red\color{black}.}
        \label{fig:swingup_visualization}
    \end{subfigure}
    \par\medskip
    \begin{subfigure}{.49\linewidth}
        \centering
        \includegraphics[width=\linewidth, height=4cm]{fig5b_student_position_swingup_study.tikz}
    \end{subfigure}
    \hfill
    \begin{subfigure}{.49\linewidth}
        \centering
        \includegraphics[width=\linewidth, height=4cm]{fig5c_student_angle_swingup_study.tikz}
    \end{subfigure}
    \par\medskip
    \vspace{-1\baselineskip}
    \begin{subfigure}{\linewidth}
        \centering
        \includegraphics[width=\linewidth, height=4cm]{fig5d_student_control_swingup_study.tikz}
        \caption{Time histories of the cartpole configurations and corresponding control inputs. The ground truth values and goal-reference trajectory are also shown.}
        \label{fig:swingup_time_histories}
    \end{subfigure}\\
    \vspace{-0.5\baselineskip}
    \caption{A successful cartpole swing up performed by a Koopman-based extension of the pixel-based linear output-feedback policy. The policy is able to overcome process noise and model mismatch to track a reference trajectory on a nonlinear system.}
    \label{fig:swingup_example}
\end{figure*}

\textbf{\begin{figure*}[h]
  \centering
  \includegraphics[width=\textwidth]{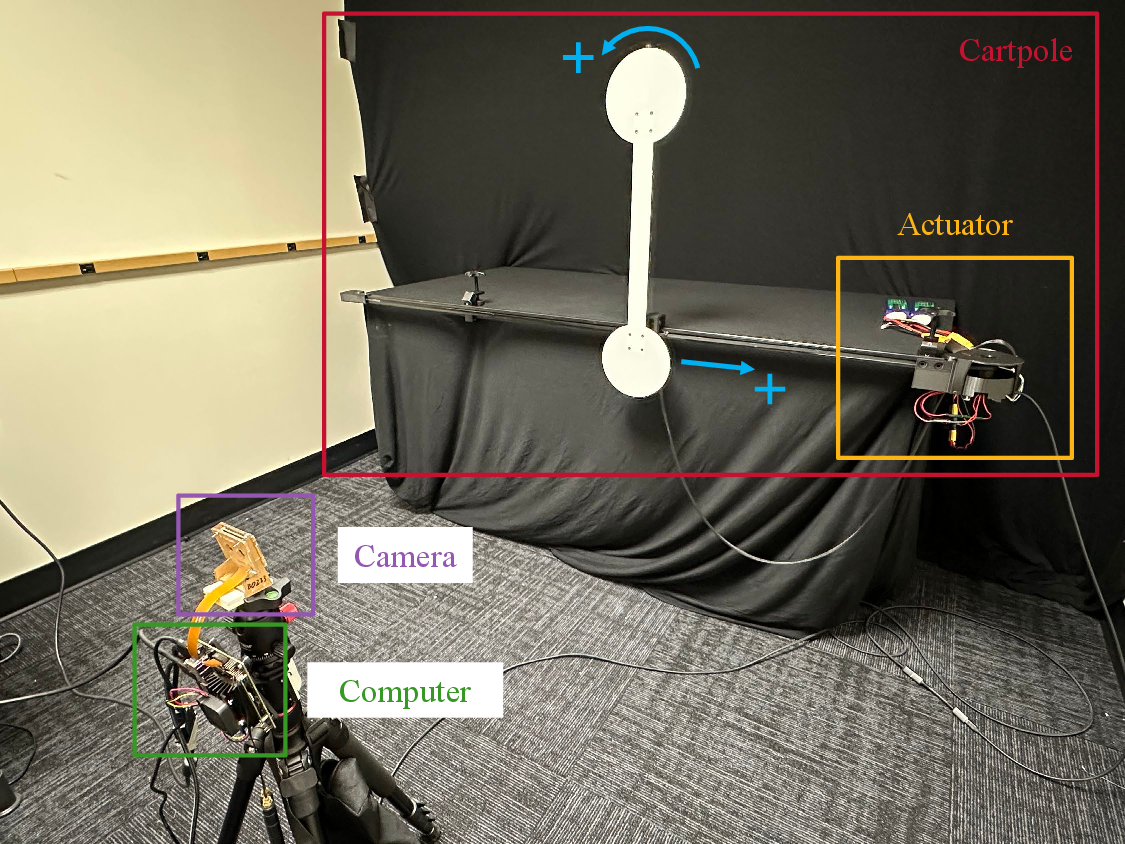}
  \caption{Overview of the cartpole hardware used for real-world demonstrations of pixels-to-torques control performed by linear output-feedback policies. The cartpole hardware is actuated by a brushless motor while an opposing global-shutter Arducam OV2311 camera captures images of the cartpole. A Jetson Nano acts as the primary computer tasked with processing the images and executing the pixel-based student policy (or corresponding teacher) before sending the control inputs to the cartpole motor. This operation is performed at 60 Hz to match the camera's maximum frame rate. Arrows indicate positive directions for the cart position, pole angle, and corresponding velocities.}
  \label{fig:pixelol_hardware_setup}
\end{figure*}}

\begin{figure*}[!htpb]
    \begin{subfigure}{\linewidth}
        \centering
        \includegraphics[width=\textwidth]{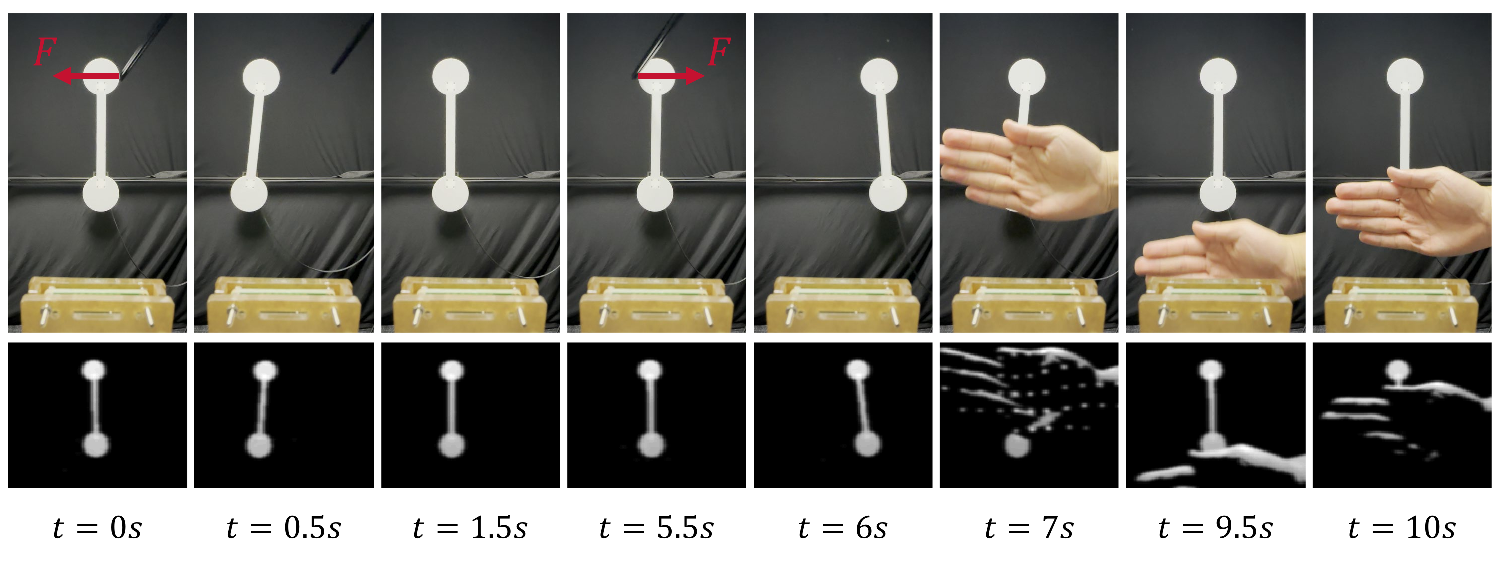}
        \caption{Snapshots of a real-world cartpole being stabilized by a pixel-based linear output-feedback policy when perturbed by a force, \color{red}$F$\color{black}, at times $t=0$ and $t=5.5$ seconds. The top row shows unfiltered images of the cartpole system taken by a separate cellphone camera while the bottom row shows the corresponding Arducam-camera images that are sent to the policy's Luenberger observer.}
        \label{fig:hardware_visualization}
    \end{subfigure}
    \par\medskip
    \begin{subfigure}{.49\linewidth}
        \centering
        \includegraphics[width=\linewidth, height=4cm]{fig7b_position_hardware_plot.tikz}
    \end{subfigure}
    \hfill
    \begin{subfigure}{.49\linewidth}
        \centering
        \includegraphics[width=\linewidth, height=4cm]{fig7c_angle_hardware_plot.tikz}
    \end{subfigure}
    \par\medskip
    \vspace{-0.5\baselineskip}
    \begin{subfigure}{\linewidth}
        \centering
        \includegraphics[width=\linewidth, height=4.75cm]{fig7d_control_hardware_plot.tikz}
        \caption{Time histories of the cartpole configurations and corresponding control inputs. The teacher's estimates and anticipated control inputs are also shown as a reference.}
        \label{fig:hardware_time_histories}
    \end{subfigure}\\
    \caption{A hardware demonstration of a pixel-based linear output-feedback policy stabilizing a real-world cartpole based on image-feedback from an Arducam OV2311 camera. Lightweight processing is performed on the camera images in the form of Gaussian blur and Otsu thresholding before being sent to the policy's pixel-based Luenberger observer. Surprisingly, the pixel-based linear output-feedback policy is able to successfully stabilize the cartpole even under the presence of large perturbations and visual occlusions.}
    \label{fig:hardware_study}
\end{figure*}

\vspace{-1\baselineskip}
\subsection{Simulation Results}

We evaluate the data-efficiency of learning the pixel-based linear output-feedback policy on cartpole-stabilization tasks as shown in Fig.~\ref{fig:stabilization_performance}. Specifically, we evaluate the policy's stabilization performance across 100 initial conditions chosen from the same distribution as the training demonstrations. Each training demonstration consists of 150 samples over 2.5 seconds. As shown in Fig.~\ref{fig:stabilization_error} and Fig.~\ref{fig:stabilization_success}, the learned pixels-to-torques policy converges to a small stabilizing error with a 97\% success rate with only 40 training trajectories. In addition, the student policy achieves a 100\% success rate when trained with only 80 trajectories. Success is defined as the cart position being within \SI{17.6}{\cm} (half the pole length) of the origin with a pole angle within \SI{2}{\degree} of upright.

As shown in Fig.~\ref{fig:observer_gain_visualization}, we also investigate the observer gain of the pixel-based linear feedback policy. Specifically, we visualize each row of the policy's observer gain, $L$, as a normalized heat map to discern the pixels that contribute most to the respective state variable's estimation. Interestingly, doing so yields identifiable visual features: Pixels that correspond to the cart contribute most to the cart-position estimate while the pole-tip pixels additionally contribute to the pole-angle estimate. In addition, the sign of the gain correctly corresponds to the physical location of the respective pixel. 

We additionally showcase the Koopman-based extension of the pixel-based linear output-feedback policy for tracking a nonlinear system as shown in Fig.~\ref{fig:swingup_example}. Specifically, we demonstrate the Koopman-based policy's ability to track a reference swing-up trajectory on the cartpole. The 76-dimensional Koopman lifting consists of the cartpole states, 4th-order Fourier features, and 6th-order Chebyshev polynomial features. The reference trajectory was designed using iterative LQR before being tracked by a time-varying-LQG teacher policy for data collection. The Koopman-based policy is trained on 20 demonstrations, each consisting of 75 samples over 2.5 seconds. As shown in Fig.~\ref{fig:swingup_time_histories}, the policy is able to properly track the swing-up trajectory despite model mismatch and process noise.

\subsection{Hardware Setup}

We develop a physical cartpole system, shown in Fig.~\ref{fig:pixelol_hardware_setup}, to evaluate the efficacy of the pixel-based linear output-feedback policy on real-world hardware. The cartpole itself consists of a single brushless motor, a cart that rides along an aluminum track, and a pole laser cut from white acrylic. A black background is added to aid image contrast, and reflective components are also covered with a black cloth during runtime. Opposing the cartpole is a Jetson Nano and Arducam OV2311 camera mounted on a tripod. The camera has a global shutter to avoid rolling-shutter distortion. The Jetson Nano acts as the main computer to perform online tasks, which consists of basic image processing (Gaussian blurring and Otsu thresholding) before executing the pixel-based linear output-feedback policy for state estimation and control. The system then sends control inputs to the motor at 60 Hz to match the frame rate of the camera. Image capture and processing is performed with OpenCV at an image resolution of $115\times140$ pixels. The pixels-to-torques student policy is trained with only 20 stabilizing demonstrations on the hardware, each with 150 samples over 2.5 seconds. The learning process is performed on a separate computer equipped with a 64-core AMD Threadripper CPU and 64GB of RAM, resulting in a solve time of about 4 minutes and 25 seconds.

\subsection{Hardware Results}\label{sec:hardware_results}

As shown in Fig.~\ref{fig:hardware_study}, the pixels-to-torques policy is able to successful recover from perturbations and stabilize the real-world cartpole. Surprisingly, the policy is also robust to major occlusions, as shown in Fig.~\ref{fig:hardware_visualization}. While the occlusions do cause decreased state-estimation performance, as shown in Fig.~\ref{fig:hardware_time_histories}, the policy's state-feedback controller is still able to overcome the disturbance to stabilize the cartpole. We attribute this to the sparse representation of $L$ as depicted in Fig.~\ref{fig:observer_gain_visualization}, which prevents a majority of the occluded pixels from affecting the policy's internal state estimate.

\section{Conclusions}\label{sec:conclusion}

We have shown that it is possible to perform “pixels-to-torques” control of a highly dynamic robotic system with simple linear output-feedback policies. These pixel-based linear policies are amenable to analysis via linear systems theory while offering surprisingly effective control in the presence of model mismatch, disturbances, and visual occlusions, as demonstrated on a real-world cartpole system. We additionally introduced a Koopman-based extension of the pixels-to-torques policy for nonlinear systems.

Our linear pixels-to-torques approach has several limitations: First, the teacher-student methodology assumes that a successful teacher policy can be designed a priori with an appropriate internal state representation. Second, the method assumes un-obstructed observations during data collection. Third, by mapping raw pixel values directly to state estimates, our approach may also be sensitive to the camera's calibration and pose. Finally, the nonlinear extension of our method also suffers from the limitations of Koopman-based approaches more broadly, which include the difficulty of crafting a good set of features or ``observables'' for lifting.

Several interesting directions for future work remain: It should be possible to apply linear visual feedback to systems with egocentric and/or multiple cameras. There are also several ways of combining our method with adaptive control or online learning techniques. Finally, we also plan to further explore the rich connection between pixel-based output feedback, Koopman operators, and diffusion policies.

\begin{credits}
\subsubsection{\ackname} This material is based upon work supported by the National Science Foundation Graduate Research Fellowship under Grant No. DGE2140739.
\end{credits}

%
%
%
\bibliographystyle{splncs04}
\bibliography{linear_pixels_to_torques}

\end{document}